# *İşletmeler için Personel Yemek Talep Miktarının Yapay Sinir Ağları Kullanılarak Tahmin Edilmesi*

# An Estimation of Personnel Food Demand Quantity for Businesses by Using Artificial Neural Networks

*Yazar(lar) (Author(s))*: M. Hanefi CALP

ORCID: 0000-0001-7991-438X

# İşletmeler için Personel Yemek Talep Miktarının Yapay Sinir Ağları Kullanılarak Tahmin Edilmesi

**M. Hanefi CALP***

Yazılım Mühendisliği Bölümü, OF Teknoloji Fakültesi, Karadeniz Teknik Üniversitesi, 61830, Trabzon, Türkiye

**ÖZ**

Günümüzde kamu veya özel kurumların birçoğu, bünyelerinde çalışan personeller için profesyonel yemek hizmeti vermektedir. Söz konusu hizmetin planlanması konusunda, kurumlarda çalışan personel sayısının genel olarak fazla olması ve personellerin şahsi veya kuruma ait sebeplerle kurum dışında olmalarından dolayı birtakım aksamalar yaşanmaktadır. Bu yüzden, günlük yemek talebinin belirlenmesi zorlaşmakta ve bu durum kurumlar için maliyet, zaman ve emek kaybına sebep olmaktadır. Bu kayıpları ortadan kaldırmak veya en azından minimuma indirmek amacıyla istatistiksel veya sezgisel yöntemler kullanılmaktadır. Bu çalışmada, işletmeler için yapay sinir ağları kullanılarak günlük yemek talebini tahmin eden yapay zekâ tabanlı bir model önerilmiştir. Veriler, günlük yemek çıkaran ve farklı kademlerde görev alan 110 kişilik bir personel kapasitesine sahip özel bir işletmenin yemekhane veritabanından elde edilmiş olup son 2 yıllık (2016-2018) veriyi kapsamaktadır. Model, MATLAB paket programı kullanılarak oluşturulmuştur. Modelin performansı, Regresyon değerleri, Ortalama Mutlak Hata Yüzdesi (OMHY-MAPE) ve Ortalama Karesel Hata (OKH-MSE) dikkate alınarak belirlenmiştir. Ağın eğitiminde, ileri beslemeli geri yayılımlı ağ mimarisi kullanılmıştır. Denemeler sonucunda elde edilen en iyi model, sırasıyla eğitim R oranı: 0,9948, test R oranı: 0,9830 ve hata oranı ise 0,003783 olup çok katmanlı (8-10-10-1) bir yapıya sahiptir. Deney sonuçları, modelin hata oranının düşük, performansının yüksek olduğunu ve talep tahmini için yapay sinir ağları kullanımının olumlu etkisini ortaya koymuştur.

**Anahtar Kelimeler: İşletme, yapay sinir ağları, yemek, talep, tahmin**

# An Estimation of Personnel Food Demand Quantity for Businesses by Using Artificial Neural Networks

**ABSTRACT**

Today, many public or private institutions provide professional food service for personnels working in their own organizations. Regarding the planning of the said service, there are some obstacles due to the fact that the number of the personnel working in the institutions is generally high and the personnel are out of the institution due to personal or institutional reasons. Because of this, it is difficult to determine the daily food demand, and this causes cost, time and labor loss for the institutions. Statistical or heuristic methods are used to remove or at least minimize these losses. In this study, an artificial intelligence model was proposed, which estimates the daily food demand quantity using artificial neural networks for businesses. The data are obtained from a refectory database of a private institution with a capacity of 110 people serving daily meals and serving at different levels, covering the last two years (2016-2018). The model was created using the MATLAB package program. The performance of the model was determinde by the Regression values, the Mean Absolute Percentage Error (MAPE) and the Mean Squared Error (MSE). In the training of the ANN model, feed forward back propagation network architecture is used. The best model obtained as a result of the experiments is a multi-layer (8-10-10-1) structure with a training R ratio of 0,9948, a testing R ratio of 0,9830 and an error rate of 0,003783, respectively. Experimental results demonstrated that the model has low error rate, high performance and positive effect of using artificial neural networks for demand estimating.

## 1. GİRİŞ (INTRODUCTION)

İşletmeler, kurum faydasını gözetmek adına çeşitli alanlarda veya konularda birtakım kararlar alabilmektedir. Bu kararlar, ya anlık ya da gelecek döneme ait olmaktadır. Kararların etkili bir şekilde alınması, işletmelerin daha kaliteli hizmet vermesi ve daha fazla kar etmesini sağlamaktadır. Bu durum, hem işletmeler hem de hizmet alanlar açısından hayati önem taşımaktadır. Bu noktada, özellikle gelecek döneme ait alınan kararlarda hata yapılması olumsuz sonuçlar doğurabilmektedir. Belirsiz durumların bulunması, işletmelerin veya kurumların geleceklerini tehlikeye atmaktadır. Bunun için önceden planlama, analiz ve tahmin faaliyetlerinin gerçekleştirilmesi gerekmektedir. İşletmelerin üretim sürecinde karar almalarını olumlu yönde etkileyen en önemli etken, üretilecek mal veya hizmetin gelecekteki satış miktarının (talebin) yüksek doğrulukta tahmin edilmesidir. Talep tahmini üzerine yapılan çalışmalarda gerçek değerler ile tahmin değerleri arasındaki fark arttıkça ihtiyaçtan daha çok üretim yapılmakta ve işletmeler yüksek oranda zarar edebilmektedir [1,2].

---

*\*Sorumlu Yazar  (Corresponding Author)*
*e-posta :  mhcalp@ktu.edu.tr*

Talep tahmini için geliştirilen yöntemlerin, basit bir algoritma veya veriye göre değişebilen özel bir model olması mümkündür. Literatürde birçok farklı sınıflandırma bulunsa da bu yöntemler nicel ve nitel olmak üzere iki başlıkta ifade edilebilir. Nicel yöntemlerde girdi için kullanılan veriler çeşitli zamanlarda gerçekleştirilen ölçümlerdir. Nitel yöntemlerde ise bu durum, uzman kişilerin alandaki beklenen gelişmeler hakkındaki fikirleridir. Aslında, her iki yöntemin de birlikte kullanıldığı durumlar vardır. Sözkonusu durumlar, nicel yöntemler kullanılarak elde edilen birtakım verilerin alandaki uzmanlar tarafından incelenmesi ile mümkündür. Bu noktada, Delphi Yöntemi, Pazar Araştırması ve Uzman Panelleri nitel yöntemlere örnek olarak verilebilir. Nicel Yöntemlere örnek olarak, Yapay Sinir Ağları, Genetik Algoritmalar, Destek Vektör Makineleri, Hareketli Ortalamalar Yöntemi, Üstel Düzleştirme Yöntemi, Ekonometrik Modeller, Basit Regresyon Analizi, Çoklu Regresyon Analizi verilebilir [3].

Bu çalışmada, işletmelerde üretilen günlük yemek miktarını veya personel talebini yapay sinir ağları kullanılarak tahmin edebilen bir model önerilmiştir. Çalışmanın amacı, işletmeler veya kurumların gelecekte üretecekleri yemek miktarlarını önceden tahmin ederek gereğinden fazla harcanacak emeğin, paranın ve zamanın önüne geçmek veya en azından bu zararları minimize etmektir. Çalışmanın diğer çalışmalardan farkı, öncelikle bu konuda yeterli sayıda çalışmanın olmamasından doğan açığı kapatması, daha sonra konuyla ilgili talep tahmin çalışmaları incelendiğinde tahmin performansının yüksek ve hata oranın da oldukça düşük olmasıdır. Ayrıca model, gerçek veriler kullanılarak daha çok çeşit menü ve modelin performansında önemli etkisi olabilecek etkenler dikkate alınarak geliştirilmiştir.

Çalışma, akış bakımından şu şekilde planlanmıştır: İkinci bölümünde, yapay sinir ağları kısaca açıklanmıştır. Üçüncü bölümde, materyal ve metot başlığı altında; verilerin elde edilmesi, hazırlanması, normalize edilmesi, eğitim ve test süreci ve modelin oluşturulması süreçleri adım adım anlatılmıştır. Dördüncü bölümde, elde edilen bulgular verilmiş ve verilen bulgular tartışılmıştır. Son olarak beşinci bölümde ise, sonuç ve öneriler başlığı altında birtakım çıkarım ve tavsiyelere yer verilmiştir.

## 2. LİTERATÜR TARAMASI (LITERATURE REVIEW)

Literatürde, günlük yemek talebinin tahmin edilmesi ile ilgili çok fazla çalışmaya rastlanmamıştır. Bu noktada, bu bölümde günlük yemek tahmini konusunda Kılıç tarafından gerçekleştirilen çalışma verildikten sonra, daha çok genel anlamda talep tahmini konusunda yapılan çalışmalara yer verilmiştir. Kılıç, Pamukkale Üniversitesi'nin yemekhane verilerini kullanarak günlük yemek miktarını yapay sinir ağları ile tahmin etmeyi amaçlayan bir model önermiştir. Önerilen sinir ağı modeli, çok katmanlı olup eğitim için Radyal Tabanlı Fonksiyon kullanılmıştır. Geliştirdiği modelle tasarlanan yemekhane günlük talep tahmin sisteminin iyi sonuç verdiği gözlenmiştir. Bu çalışma ile, diğer talep tahminleri gibi yemekhane talep tahmininde de yapay sinir ağları ve diğer yöntemlerin başarılı bir şekilde kullanılabileceğini ortaya koymuştur [1].

Köksal ve Uğursal, konut kullanımındaki enerji tüketimini modellemek için koşullu talep analizi (KTA) yönteminin kullanımını araştırmışlardır. KTA'nın bölgesel düzeyde enerji tüketimini modellemek için kullanıldığı birkaç çalışma olduğunu, ancak KTA yönteminin, ulusal düzeyde konut enerji tüketimini modellemek için kullanılmadığını belirtmişlerdir. Tahmin performansı ve KTA modelinin konut kullanımındaki enerji tüketimini karakterize etme yeteneği, daha önce geliştirilen model ve bir sinir ağı ile karşılaştırılmıştır. Modellerin tahminlerinin karşılaştırılması, KTA'nın konut sektöründeki enerji tüketimini ve diğer iki modeli doğru bir şekilde tahmin edebildiğini göstermektedir. Sonuç olarak, sınırlı sayıda değişkene bağlı olarak, KTA modelinin değerlendirebilme kabiliyetinin sinir ağı modelinden daha düşük olduğunu bulmuşlardır [4].

Murat ve Ceylan, sosyo-ekonomik ve ulaşımla ilgili göstergeleri kullanarak ulaşım enerji talebi tahmini için denetimli sinir ağlarına dayalı yapay sinir ağı (YSA) yaklaşımı önermişlerdir. Kısaca, YSA ulaşım enerjisi talebi modeli geliştirmişlerdir. Tahmin işleminde, ileri beslemeli geri yayılımlı bir sinir ağı kullanılmışlardır. YSA, Sosyo-ekonomik göstergelerin ulaştırma enerji talebi üzerindeki etkisini araştırmak için 1970'den 2001'e kadar olan mevcut enerji verileriyle birlikte, gayri safi milli hasıla, nüfus ve toplam yıllık ortalama araç kilometresine göre analiz edilmiştir. Test süresindeki enerji verileri ile model tahminleri karşılaştırılarak model yüksek doğruluğunu ortaya koymuşlardır. Test iki senaryo ile yapılmıştır. YSA'nın hem bağımlı hem de bağımsız değişkenler için tarihsel verilerdeki dalgalanmaları yansıttığı anlaşılmaktadır. Elde edilen sonuçlarla, ulaştırma enerji tahmini problemi için benimsenen metodolojinin uygunluğunu ortaya koymuşlardır [5].

Geem ve Roper, Güney Kore için enerji talebini etkin bir şekilde tahmin etmek için yapay bir sinir ağı modeli önermişlerdir. Sözkonusu model, ileri beslemeli çok katmanlı algılayıcı, hata geri yayılım algoritması, momentum süreci ve ölçeklenmiş veriler içeren bir yapıya sahiptir. Model, gayri safi yurtiçi hasıla, ithalat ve ihracat tutarları ve nüfus gibi dört bağımsız değişken içerir. Veriler, çeşitli yerel ve uluslararası kaynaklardan elde edilmiştir. Önerilen modelin, herhangi bir aşırı uyum sorunu olmaksızın doğrusal bir regresyon modelinden veya üstel modelden (kök ortalama karesi hatası (RMSE) açısından) daha iyi tahmin ettiği belirtmişlerdir. Sonuçlar, kalıcı olarak büyümek yerine, enerji talepleri belirli noktalarda zirveye çıktığını ve sonra yavaş yavaş azaldığını, bu eğilimin, regresyon veya üstel model ile elde edilen sonuçlardan oldukça farklı olduğunu göstermiştir [6].

Doğan ve diğ., su kaynaklarının değişkenlerini tahmin etmek için YSA'nın giderek daha fazla kullanıldığını ve ileri beslemeli nöral ağ modelleme tekniği, su kaynakları uygulamalarında en çok kullanılan YSA tipi olduğunu belirterek çalışmalarında biyolojik oksijen talebi (BOT) tahmininin doğruluğunu geliştirmek için YSA modelinin yeteneklerini araştırmışlardır. Bunun için 2001 yılında Melen Havzası'ndaki 11 örnekleme sahasından veriler toplanmıştır. BOT'u tahmin etmek ve bir YSA modeli geliştirmek için mevcut veri seti bir eğitim setine ve bir test setine ayrılmıştır. En uygun miktarda gizli katman düğümüne ulaşmak için, 2, 3, 5, 10 nodları test edilmiştir. Bu aralıkta, 8 girişli ve 3 düğümlü 1 gizli katmanı olan YSA mimarisi en yüksek performansa sahip olarak elde edilmiştir. Sonuçların karşılaştırması, YSA modelinin BOT tahmini için makul tahminler verdiğini ortaya koymuştur [7].

Karahan, doktora tezinde Malatya şehrinde üretilen kuru kayısı meyvesinin talep miktarını çok katmanlı yapay sinir ağı kullanarak tahmin etme üzerine bir çalışma yapmıştır. Kuru kayısı meyvesinin talep miktarını doğrudan etkileyecek değişkenleri geçmiş talep miktarı bilgileri, ABD doları kur bilgisi (aylık), ortalama ürün satış fiyatı (aylık), satış yapılan pazar sayısı ve mevsimsel faktörler olarak belirlemiştir. Bu faktörler, aynı zamanda modelin girişlerini oluşturmaktadır. Önerilen modelin çıkışı ise, belirlenen tarihteki kuru kayısı talep miktarıdır. Model için 2004-2011 yılları arasındaki toplam 84 aylık veriyi almıştır. Eğitim için 70 ve test için kalan 14 aylık veri miktarını rastgele seçerek çalışmada kullanmıştır. Çalışmanın performansını test etmek için hata ölçüm fonksiyonu olan RMSE kullanmıştır. Modeli kullanarak 2011 yılının ilk 6 ayı talep miktarını tahmin etmiştir. Sonuçlarla, önerdiği modelin performans kalitesinin yüksek olduğunu ortaya koymuştur [8].

Sönmez ve diğ., Türkiye'nin ulaştırma enerji talebini tahmin etmek için yapay arı kolonisi algoritmasını kullanarak üç farklı matematiksel model önermişlerdir. Parametre olarak, gayri safi yurtiçi hasıla, nüfus ve toplam yıllık araç-km değişkenleri alınmıştır. Ulaşım enerji talebi tahminleri için doğrusal, üstel ve ikinci dereceden matematiksel ifadeler kullanılmıştır. Eğitim ve test aşamaları için 1970-2013 arasındaki 44 yıllık tarihsel verilerden yararlanılmıştır. Modellerin performansları altı farklı global hata ölçüm yaklaşımıyla değerlendirilmiştir. Geliştirilen modeller, Türkiye'nin ulaştırma enerji talebini 2014'ten 2034'e kadar olan 21 yıllık bir dönemde tahmin etmek için iki olası senaryoda kullanılmıştır. Yapay arı kolonisi algoritması, Türkiye'de ulaştırma enerji planlaması ve politika gelişmeleri için optimizasyon yönteminin uygunluğunu ortaya koymuştur. Ayrıca, senaryolardan elde edilen sonuçlar, Türkiye'nin enerji talebinin 2034 yılına kadar 2013 yılının iki katı olacağını göstermiştir [9].

Zeng ve diğ., çalışmada, enerji tüketimini tahmin etmek için uyarlamalı diferansiyel evrim algoritması tarafından desteklenen geri yayılımlı sinir ağı modeli olan ADE-BPNN adlı hibrit bir akıllı yaklaşımın uygulanmasını amaçlamışlardır. Önerdikleri hibrit model, gayri safi yurtiçi hasıla, nüfus, ithalat ve ihracat verilerini girdi olarak kullanmaktadır. Adaptif mutasyon ve çaprazlama ile geliştirilmiş bir diferansiyel evrim, BPNN'nin tahmin performansını artırmak için uygun küresel başlangıç bağlantı ağırlıklarını ve eşikleri bulmak için kullanılmıştır. Önerilen ADE-BPNN modelinin uygulanabilirliği ve doğruluğunu ortaya koymak için karşılaştırmalı bir örnek ve iki genişletilmiş örnek kullanılmıştır. Test veri setlerinin hataları, ADE-BPNN modelinin geleneksel geri yayılımlı sinir ağı modeli ve diğer popüler mevcut modellerle karşılaştırıldığında enerji tüketimini etkili bir şekilde tahmin edebildiğini göstermektedir. Ayrıca, ABD'deki elektrik enerjisi tüketimi ve Çin'deki toplam enerji tüketimi tahmini ve etkili enerji tüketimi tahmininin iyileştirilmesi için her bir girdi değişkeninin göreceli önemini niceliksel olarak araştırmak üzere ortalama etki değeri bazlı analiz gerçekleştirilmiştir [10].

Moradi ve diğ., doğrusal olmayan regresyon analizi yöntemi ile kloramin bozulma oranını etkileyen, kimyasal ve mikrobiyolojik faktörlere dayalı olarak tam ölçekli içme suyu kaynaklarında kloramin talebinin tahmin edilmesine olanak sağlayan bir model geliştirmişlerdir. Model, su örneklerinin organik karakterine (spesifik ultraviyole absorbansı (SUVA)) ve kloraminin mikrobiyolojik (Fm) çürümesine dair bir laboratuvar ölçümüne dayanmaktadır. Kloramin artığının tahmin edilmesi için modelin uygulanabilirliği (ve dolayısıyla kloramin talebi), Avustralya'daki farklı su arıtma tesislerinden gelen çeşitli sularda deneysel ve tahmin edilen veriler arasındaki istatistiksel test analizi ile test edilmiştir. Sonuçlar, modelin gerçek içme suyu sistemlerinde çeşitli zamanlarda kloramin talebini simüle edebildiğini ve tahmin edebildiğini göstermiştir. Avustralya'da üç su kaynağı için modelin kinetik parametreleri olarak tahmin edilen hızlı ve yavaş bozulma oranı sabitlerinin önemi tartışılmıştır. Aynı su kaynağı ile kinetik parametrelerin aynı kaldığı bulunmuştur. Ayrıca, önerilen modelleme yaklaşımının, kloramin dezenfeksiyonunu yönetmek için su arıtma operatörleri tarafından bir karar destek aracı olarak kullanılma potansiyeline sahip olduğunu belirtmişlerdir [11].

Ahmed, yaptığı çalışmada, ileri beslemeli bir sinir ağ modeli ve Bangladeş Surma nehrinde biyokimyasal oksijen talebinden ve kimyasal oksijen talebinden çözünmüş oksijeni öngören radyal temel işlevli sinir ağı modeli geliştirmeyi amaçlamıştır. Sinir ağı modeli, üç yıllık bir çalışma sırasında toplanan deneysel veriler kullanılarak geliştirilmiştir. Giriş kombinasyonları çözünmüş oksijen ile korelasyon katsayısına göre hazırlanmıştır. YSA modellerinin performansı korelasyon katsayısı (R), ortalama karesel hata (MSE) ve etkinlik katsayısı (E) kullanılarak değerlendirilmiştir. Surma nehrinin çözünmüş oksijeninin tahmininde YSA modelinin başarılı bir şekilde kullanılabileceği bulunmuştur. Optimal RBFNN'nin biyokimyasal oksijen ihtiyacı ve kimyasal oksijen ihtiyacı ile test dizisi tahminleri için MSE = 0.465, E = 0.905 ve R = 0.904 elde edilmiştir. RBFNN ve FFNN ile modellenen değerler deneysel verilerle karşılaştırılarak önerilen sinir ağı modelinin makul sonuçlar verdiğini göstermiştir [12].

Ay ve Kişi, çalışmalarında, çözünmüş oksijen konsantrasyonunu modellemek için çok katmanlı bir perceptron, radyal temel nöral ağ ve iki farklı uyarlamalı nöro-bulanık çıkarım sistemi yönteminde yer alan ileri kemometrik teknikler geliştirmişlerdir. Ayrıca bu modellerin tahminlerini çoklu doğrusal regresyon ile karşılaştırmışlardır. Bu bağlamda, ABD'de SC Carlisle yakınlarındaki Broad River'da kaydedilen sıcaklık, pH, elektrik iletkenliği, deşarj ve çözünmüş oksijen verilerinin aylık ortalama miktarları kullanılmıştır. Modellerin doğruluğu, determinasyon katsayısı, MSE, MAPE ve RMSE kullanılarak bir diğeri ile karşılaştırılmıştır. Sonuçlar, radyal temelli sinir ağı yönteminin, aylık ortalama çözünmüş oksijen konsantrasyonunu modellemede diğer yöntemlerden daha iyi performans gösterdiğini göstermiştir. Ayrıca, sıcaklık, pH, elektrik iletkenliği ve deşarjın çözünmüş oksijen konsantrasyonu üzerinde etkili olduğu bulunmuştur [13].

Khoshravesh, çalışmada, aylık referans evapotranspirasyonu, Ardestan, Esfahan ve Kashan'da çok değişkenli fraksiyonel polinom (MFP), sağlam regresyon ve Bayes regresyon gibi üç farklı regresyon modeli ile tahmin edilmiştir. Sonuçlar en iyi modeli seçmek için Gıda ve Tarım Örgütü (FAO) -Penman – Monteith (FAO-PM) ile karşılaştırılmıştır. Sonuçlar, aylık olarak tüm modellerin FAO-PM ($R^2 > 0.95$ ve RMSE < 12.07 mm ay-1) için hesaplanan değerler ile daha yakın bir anlaşma sağladığını göstermiştir. Bununla birlikte, MFP modeli, tüm istasyonlarda referans evapotranspirasyon tahmini için diğer iki modelden daha iyi tahminler verdiği ortaya çıkmıştır [14].

## 3. MATERYAL VE METOT (MATERIAL AND METHOD)

Çalışmanın bu bölümünde, öncelikle çalışmada kullanılan YSA konusu kısaca açıklanmıştır. Daha sonra sırasıyla, verilerin elde edilmesi, elde edilen verilerin sayısallaştırılması, veriler içerisinden gereksiz veya normalden çok uzak olan verilerin temizlenmesi (gürültüden arındırma), gürültüden arındırılmış verilerin normalize edilmesi ve son olarak tüm veriler belirlenen oranda eğitim ve test olarak ikiye ayrılarak modelin oluşturulması aşamaları adım adım tüm ayrıntılarıyla verilmiştir.

### 3.1. Yapay Sinir Ağları (Artificial Neural Networks)

YSA, biyolojik sinir hücresi (nöron) yapısı dikkate alınarak modellenen ve süreç içerisinde kendi kendine öğrenme yeteneğine sahip bir algoritmadır. YSA, robot teknolojisi, desen tanıma, tıp, güç sistemleri, sinyal işleme, tahmin ve özellikle sistem modelleme gibi birçok farklı uygulamalarda kullanılabilmektedir [15-22]. Basit bir YSA örneği Şekil 1'de gösterilmiştir.

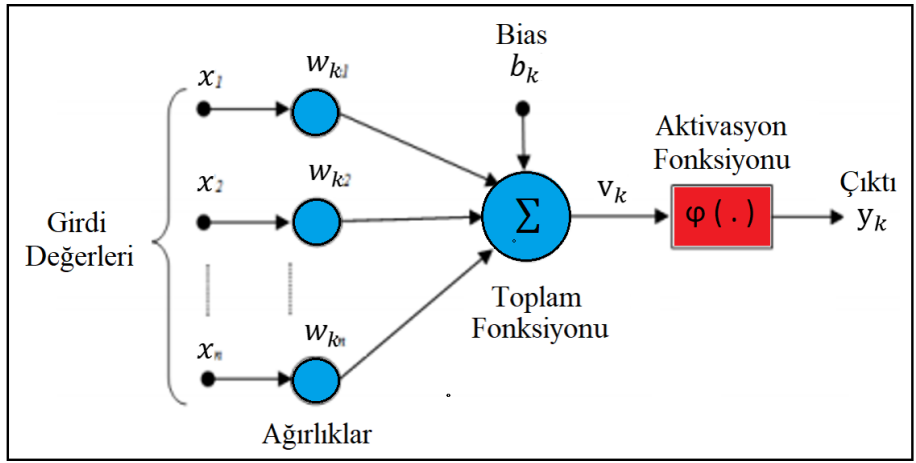

**Şekil 1.** Basit bir YSA örneği (A simple ANN) [23,24]

Şekil 1'de görülen YSA modeli, matematiksel olarak Eşitlik 1'deki gibi tanımlanabilir.

$$u_k = \sum_{j=1}^{n} w_{kj} x_j$$
$$y_k = \varphi(u_k + b_k)$$
$$v_k = u_k + b_k$$
$$y_k = \varphi(v_k) \qquad (1)$$

YSA, nöronların birbirleriyle bağlantılar aracılığıyla bir araya gelmelerinden oluşmaktadır [23,25]. Ayrıca, kendisine verilen çeşitli örnek verilerden çıkarım veya genelleme yaparak öğrenir ve böylece yeni bilgiler türetebilir. YSA tekniği, doğrusal olmayan problemlerin çözümünde kullanılır. Veriler, eğitim ve test kümesi olarak ikiye bölünür. Eğitim sürecinin amacı, sinir ağındaki ağırlıkları ayarlayarak hata düzeyini düşürmek veya minimize etmektir. Bu süreç, amaçlanan çıktı elde edilinceye kadar devam eder. Eğitim işleminin performans düzeyi, eğitim esnasında kullanılmayan verilerin sinir ağında test edilmesiyle elde edilir [26-30]. Ağın eğitiminde, ileri beslemeli geri yayılımlı sinir ağı mimarisi etkilidir. İleri beslemeli sinir ağı, modelin girdi katmanından çıktı katmanına doğru tek yönde ilerlemektedir. İleri beslemeli YSA modeli, öncelikle girdi katmanı, daha sonra bir veya iki gizli katman ve son olarak çıktı katmanından oluşmaktadır (Şekil 2) [31-33].

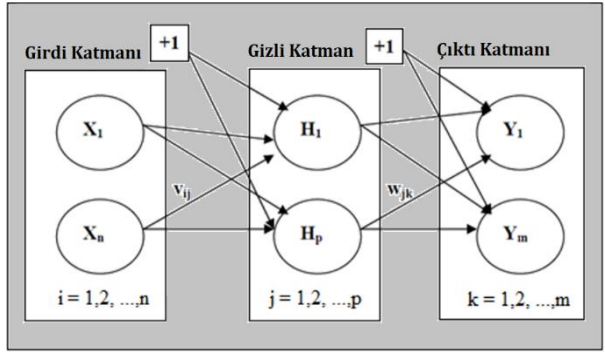

**Şekil 2.** İleri beslemeli YSA (Feed forward ANN)

Ağın eğitimi, her bir katmanda yer alan nöronlar arasındaki bağlantıların ağırlıklarının düzenlenmesi ile gerçekleştirilmektedir. Bu işlem, Eşitlik 2'de sunulan hata fonksiyonu ile gerçekleştirilir [33,34] ($d_j$: hedeflenen sonuç, $o_j$: gerçekleşen sonuç).

$$E^p = \frac{1}{2}\sum_j (d_j^p - o_j^p)^2 \quad (2)$$

Eşitlik 3'te verilen hata fonksiyonunun farkı, ağırlıkların yeniden düzenlenmesi için kullanılmaktadır.

$$\Delta^p w_{ji} = -\eta \cdot \left(\frac{\partial E^p}{\partial w_{ji}}\right) \quad (3)$$

Bu denklemde, η sabiti (öğrenme oranı) için herhangi bir değer atanabilir. Ağırlıkların yeniden düzenlenmesi için Eşitlik 4 kullanılır.

$$w_{ij}(t+1) \cong w_{ij}(t) + \eta \cdot \delta_j \cdot i_i \quad (4)$$

Bu denklemde, $w_{ij}(t)$: ağırlık, $i_i$: i düğümünün sonuç değeri veya $\delta_j$: j düğümünün hata terimidir. Çıktı katmanındaki düğüm için hata: ($\delta_j$);

$$\delta_j \cong o_j \cdot (1 - o_j)(d_j - o_j) \quad (5)$$

formülüyle hesaplanır. $j$ düğümü bir gizli düğüm olmak üzere hata terimi ($\delta_j$) (Eşitlik 6);

$$\delta_j \cong o_j \cdot (1 - o_j) \cdot \sum_k \delta_k \cdot w_{jk} \quad (6)$$

Ağırlık değişimleri, herhangi bir "moment" teriminin (α) eklenmesi ile revize edilebilir.

$$w_{ij}(t+1) \cong w_{ij}(t) + \eta \cdot (d_j - o_j) \cdot i_i + \alpha \cdot (w_{ij}(t) - w_{ij}(t-1)) \quad (7)$$

**3.2. Verilerin Elde Edilmesi ve Hazırlanması** (Obtaining and Preparing of the Data)

Veriler, günlük yemek çıkaran ve farklı kademelerde görev alan 110 kişilik bir personel kapasitesine sahip özel bir işletmenin yemekhane veritabanından elde edilmiş olup son 2 yıllık (2016-2018) toplam 730 satır veriyi kapsamaktadır. Ancak, bu veriler içerisinde normalden çok uzak veya eksik veri olan kayıtlar (toplam 180 satır) bulunduğu için modelin oluşturulmasında geriye kalan 550 satırlık veri kullanılmıştır. İşletme adı, kurum mahremiyeti açısından çalışma içerisinde "ABC" olarak verilmiştir.

**3.3. Modelin Oluşturulması** (Creating of the Model)

Modeli oluşturmadan önce, girişleri oluşturan başlıklar belirlenmiştir. Yemek hizmeti veren ABC işletmesinin günlük menüsünde Çorba, Ana Yemek, Ek Yemek, Ek Yardımcı Yemek ve İçecek olmak üzere beş (5) çeşit yemek bulunmaktadır. Menüde yer alan her bir yemeğin adı ve sayısı Çizelge 1'de verilmiştir.

**Çizelge 1.** İşletmedeki menüler ve çeşit sayıları (Menus in the business and number of kinds)

| Sıra No | Ana Menü Adı | Alt Menü Adı | Çeşit |
|---|---|---|---|
| 1 | Çorba | Mercimek Çorbası, Ezogelin Çorbası, Domates Çorbası, Yayla Çorbası, Şehriye Çorbası, Tarhana Çorbası, Erişte Çorbası, Ayran Aşı Çorbası, Brokoli Çorbası, | 9 |
| 2 | Ana Yemek | Kuru Fasulye, Bamya, Sulu Köfte, Sebze Dolması, Mantı, Türlü, Patlıcan Musakka, Tavuk Pirzola, Çiftlik Kebabı, Kadın Budu Köfte Patates, Püreli Dana Rosto, Güveç, Ispanak, Balık | 14 |
| 3 | Ek Yemek | Pirinç Pilavı, Bulgur Pilavı, Makarna, Su Böreği | 4 |
| 4 | Ek Yardımcı Yemek | Salata, Tatlı, Turşu, Meyve | 4 |
| 5 | İçecek | Su, Ayran, Kola, Soda, | 4 |

Model oluşturulurken günlük talebi ve tahmin sonucunu etkileyeceği düşünülen faktörler özellikle dikkate alınmıştır. Bu faktörler; Çorba, Ana Yemek, Ek Yemek, Ek Yardımcı Yemek, İçecek, Haftanın Günü (Pazartesi, Salı, Çarşamba, Perşembe, Cuma, Cumartesi, Pazar), Resmi Tatil (Var, Yok) ve Mevsim (Sonbahar, Kış, İlkbahar, Yaz) olarak belirlenmiştir. Böylece, model sekiz (8) girişten oluşmaktadır.

*Verilerin normalize edilmesi (Normalizing of the data)*
YSA modeli oluşturulmadan önce, gereksiz veri tekrarını önlemek, verileri 0-1 bandında sınırlamak ve böylece oluşturulacak olan YSA modelinin performansını arttırmak amacıyla veriler normalize edilmiştir. Bu çalışmada, elde edilen veriler, öncelikle sayısallaştırılmıştır (Çizelge 2).

**Çizelge 2.** Menülerin sayısallaştırılması ve normalize değerleri (Digitalization of menus and their normalized values)

| Sıra No | Ana Menü Adı | Alt Menü Adı | Değer | Normalize Edilmiş Değer |
|---|---|---|---|---|
| 1 | Çorba | Mercimek Çorbası | 1 | 0 |
| | | Ezogelin Çorbası | 2 | 0,125 |
| | | Domates Çorbası | 3 | 0,25 |
| | | Yayla Çorbası | 4 | 0,375 |
| | | Şehriye Çorbası | 5 | 0,5 |
| | | Tarhana Çorbası | 6 | 0,625 |
| | | Erişte Çorbası | 7 | 0,75 |
| | | Ayran Aşı Çorbası | 8 | 0,875 |
| | | Brokoli Çorbası | 9 | 1 |
| 2 | Ana Yemek | Kuru Fasulye | 1 | 0 |
| | | Bamya | 2 | 0,076923 |
| | | Sulu Köfte | 3 | 0,153846 |
| | | Sebze Dolması | 4 | 0,230769 |
| | | Mantı | 5 | 0,307692 |
| | | Türlü, | 6 | 0,384615 |
| | | Patlıcan Musakka | 7 | 0,461538 |
| | | Tavuk Pirzola | 8 | 0,538462 |
| | | Çiftlik Kebabı | 9 | 0,615385 |
| | | Kadın Budu Köfte Patates | 10 | 0,692308 |
| | | Püreli Dana Rosto | 11 | 0,769231 |
| | | Güveç | 12 | 0,846154 |
| | | Ispanak | 13 | 0,923077 |
| | | Balık | 14 | 1 |
| 3 | Ek Yemek | Pirinç Pilavı | 1 | 0 |
| | | Bulgur Pilavı | 2 | 0,333333 |
| | | Makarna | 3 | 0,666667 |
| | | Su Böreği | 4 | 1 |
| 4 | Ek Yardımcı Yemek | Salata | 1 | 0 |
| | | Tatlı | 2 | 0,333333 |
| | | Turşu | 3 | 0,666667 |
| | | Meyve | 4 | 1 |
| 5 | İçecek | Su | 1 | 0 |
| | | Ayran | 2 | 0,333333 |
| | | Kola | 3 | 0,666667 |
| | | Soda | 4 | 1 |

Çizelge 3'te verilen değerler, Çizelge 2'deki gibi herbir alt menünün sırasıyla numaralandırılması (sayısallaştırılması) ve daha sonra bu değerlerin normalize edilmesi sonucu elde edilmiştir. Bu noktada, normalize sürecinde min-max yöntemi kullanılmıştır. Burada, $v_R$ girdinin gerçek değerini, $v_{min}$ minimum girdi değerini, $v_{max}$ ise maksimum girdi değerini ifade etmektedir (Eşitlik 8) [19,33,34].

$$V_n = \frac{V_R - V_{min}}{V_{max} - V_{min}} \quad (8)$$

Normalize edilen eğitim ve test verilerinin bir kısmı Çizelge 3'te verilmiştir.

**Çizelge 3.** Verilerin bir kısmı (Part of the data)

| Veri Tipi | Çorba | Ana Yemek | Ek Yemek | Ek Yardımcı Yemek | İçecek | Haftanın Günü | Resmi Tatil | Mevsim | Talep Miktarı |
|---|---|---|---|---|---|---|---|---|---|
| Eğitim Verisi | 0,0000 | 0,1538 | 0,3333 | 0,0000 | 0,3333 | 0,5000 | 1,0000 | 0,6667 | 0,7222 |
| | 0,0000 | 0,0769 | 0,3333 | 0,6667 | 0,6667 | 0,8333 | 0,0000 | 0,0000 | 0,0000 |
| | 0,1250 | 0,5385 | 0,6667 | 0,3333 | 0,0000 | 1,0000 | 1,0000 | 0,3333 | 0,6111 |
| | 0,5000 | 0,4615 | 1,0000 | 0,3333 | 0,0000 | 0,0000 | 1,0000 | 1,0000 | 0,9444 |
| | 1,0000 | 0,2308 | 0,3333 | 0,0000 | 1,0000 | 0,0000 | 1,0000 | 0,6667 | 0,9074 |
| | 0,0000 | 0,3846 | 0,0000 | 0,0000 | 0,3333 | 0,5000 | 1,0000 | 0,6667 | 0,7037 |
| | 0,1250 | 0,6154 | 0,6667 | 0,3333 | 0,3333 | 0,1667 | 1,0000 | 0,3333 | 0,5463 |
| | 0,2500 | 0,7692 | 0,6667 | 0,6667 | 0,3333 | 0,1667 | 0,0000 | 0,6667 | 0,0000 |
| | 0,5000 | 0,3077 | 0,6667 | 0,3333 | 0,3333 | 0,1667 | 1,0000 | 0,0000 | 0,8889 |
| | 0,6250 | 0,5385 | 1,0000 | 1,0000 | 1,0000 | 0,5000 | 1,0000 | 1,0000 | 0,8056 |
| Test Verisi | 0,8750 | 0,4615 | 0,0000 | 0,3333 | 1,0000 | 0,1667 | 0,0000 | 0,3333 | 0,0000 |
| | 0,0000 | 0,8462 | 1,0000 | 0,3333 | 0,3333 | 0,5000 | 1,0000 | 0,6667 | 0,9167 |
| | 0,3750 | 0,0000 | 0,3333 | 0,3333 | 1,0000 | 0,3333 | 1,0000 | 0,6667 | 0,7963 |
| | 0,5000 | 0,0000 | 0,6667 | 0,3333 | 0,3333 | 0,0000 | 1,0000 | 0,3333 | 0,7222 |
| | 1,0000 | 0,9231 | 0,6667 | 0,3333 | 0,3333 | 0,0000 | 1,0000 | 0,3333 | 0,7407 |
| | 0,8750 | 0,6923 | 0,0000 | 0,6667 | 0,3333 | 0,5000 | 1,0000 | 0,6667 | 0,8889 |
| | 0,5000 | 0,6154 | 0,3333 | 0,0000 | 0,3333 | 0,1667 | 1,0000 | 1,0000 | 0,7685 |
| | 0,0000 | 0,5385 | 0,3333 | 0,6667 | 1,0000 | 0,3333 | 1,0000 | 0,6667 | 0,9259 |
| | 0,7500 | 0,3846 | 0,0000 | 0,3333 | 0,6667 | 0,0000 | 1,0000 | 0,0000 | 0,8333 |
| | 0,5000 | 0,0000 | 0,6667 | 0,3333 | 0,3333 | 0,3333 | 1,0000 | 1,0000 | 0,7963 |

***Eğitim ve test süreci*** *(The training and testing process)*
Veriler normalize edildikten sonra eğitim ve test işlemine geçilmiştir. Bu işlemler, MATLAB paket programı kullanılarak gerçekleştirilmiştir. Gürültülü veriler temizlendikten sonra elde edilen toplam 550 adet gerçek veri içerisinden rastgele belirlenen %70'lik kısmı eğitim, %30'luk kısmı ise test için ayrılmıştır. Daha sonra eğitim ve test süreçleri uygulanmıştır. Önerilen modeli oluşturmak için farklı şekillerde birçok deneme yapılmıştır. Denemeler sonucunda, en iyi model (yüksek R ve düşük hata oranı) belirlenmiştir. Sözkonusu denemelerden elde edilen bulguların bir kısmı Çizelge 4'te verilmiştir.

**Çizelge 4.** Farklı modeller ve performans sonuçları (Different models and performance results)

| No | Model | Eğitim Fonksiyonu | Aktivasyon Fonksiyonu | Gizli Katman Sayısı | Gizli Katmandaki Nöron Sayısı | Performans Kriterleri | | |
|---|---|---|---|---|---|---|---|---|
| | | | | | | Eğitim R Oranı | Test R Oranı | Ortalama Karesel Hata |
| 1 | 8-5-1 | trainlm | Logsig-Tansig | 1 | 5 | 0,9675 | 0,8414 | 0,06 |
| 2 | 8-5-1 | trainlm | Tansig- Tansig | 1 | 5 | 0,9681 | 0,8580 | 0,05 |
| 3 | 8-5-5-1 | trainlm | Logsig-Logsig-Tansig | 2 | 5-5 | 0,9789 | 0,9581 | 0,04 |
| 4 | 8-5-5-1 | trainlm | Logsig-Tansig Tansig | 2 | 5-5 | 0,9807 | 0,9516 | 0,03 |
| 5 | 8-10-5-1 | trainlm | Logsig-Logsig-Tansig | 2 | 10-5 | 0,9885 | 0,9722 | 0,02 |
| 6 | 8-10-5-1 | trainlm | Logsig-Tansig Tansig | 2 | 10-5 | 0,9847 | 0,9686 | 0,02 |
| 7 | 8-10-10-1 | trainlm | Logsig-Logsig-Tansig | 2 | 10-10 | 0,9948 | 0,9830 | 0,003 |
| 8 | 8-10-10-1 | trainlm | Logsig-Tansig Tansig | 2 | 10-10 | 0,9890 | 0,9858 | 0,004 |
| 9 | 8-10-15-1 | trainlm | Logsig-Logsig-Tansig | 1 | 5 | 0,9745 | 0,9601 | 0,04 |
| 10 | 8-10-15-1 | trainlm | Logsig-Tansig Tansig | 1 | 5 | 0,9708 | 0,9658 | 0,03 |

Çizelge 4 incelendiğinde, model tek gizli katmanlı iken oldukça düşük performansa sahip olduğu, ancak iki gizli katmana sahip olduğunda tam tersi hata oranının düştüğü görülmektedir. Buna ilaveten, iki gizli katman olmasına rağmen gizli katmanda yer alan nöron sayısı arttırıldığında da sinir ağı modelinin performansının önemli ölçüde düştüğü görülmektedir. Bu durum, modelin ya ezberlediğine veya öğrendiğini unuttuğuna işarettir.

Veriler eğitildikten sonra yapılan denemeler sonucunda en iyi model; eğitim R oranı: 0,9948, test R oranı: 0,9830 ve hata oranı ise 0,003783 olup çok katmanlı bir yapıya (8-10-10-1) sahiptir. Elde edilen değerler, modelin ezberleme durumunun ve performans düzeyinin ölçülmesi için test verileri ile test edilmiştir. Belirlenen modelin eğitim–test R değerleri ve hata grafiği Şekil 3'te verilmiştir.

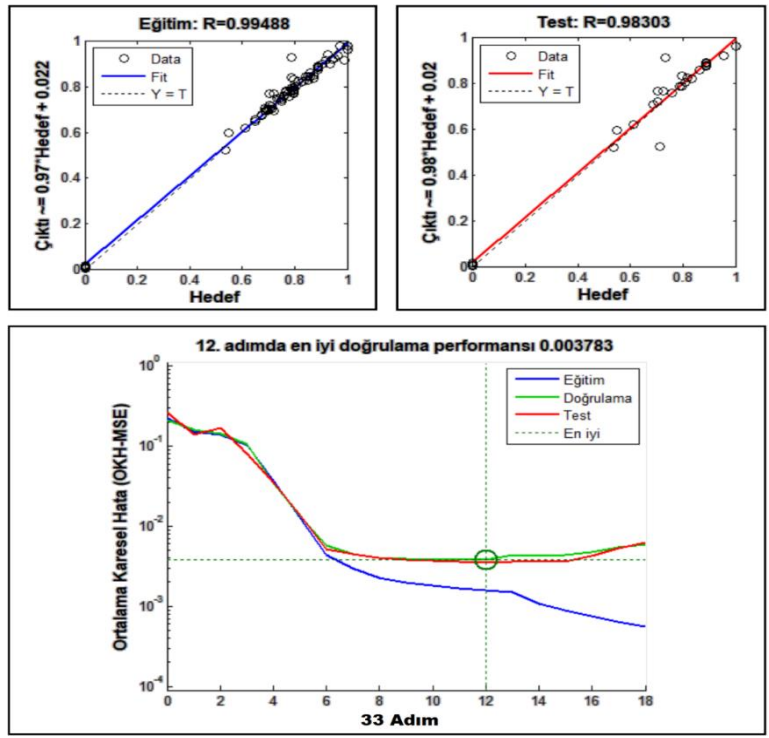

**Şekil 3.** Modelin eğitim ve test R değerleri ve hata grafiği (Training and testing R values and error graph of the model)

Çok katmanlı bir yapıya sahip olan YSA modeli Şekil 4'te verilmiştir.

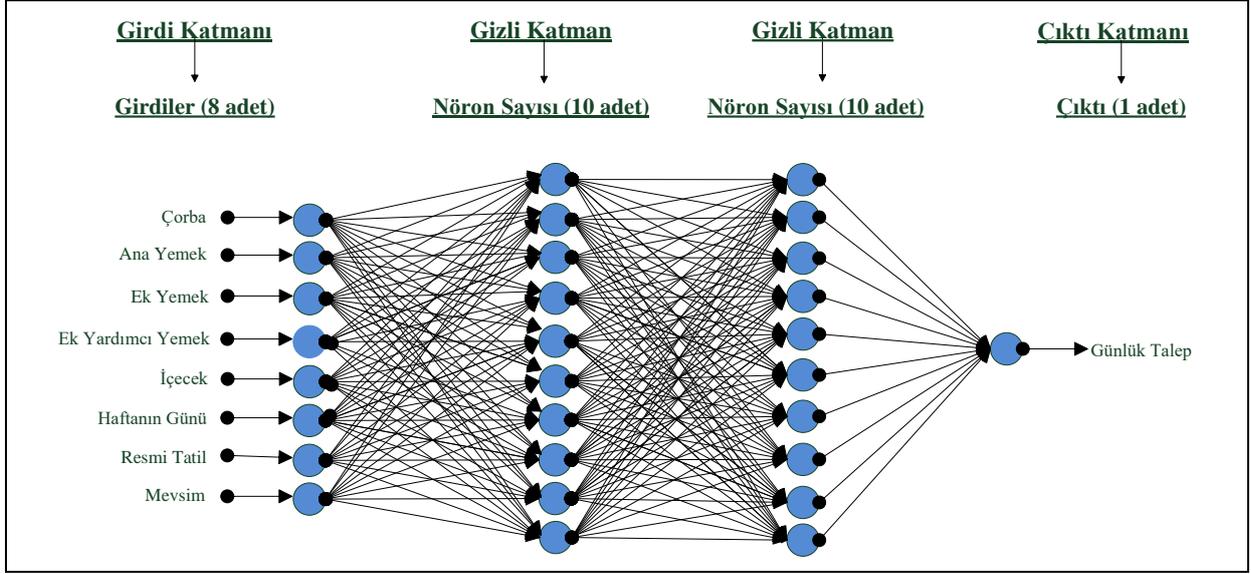

**Şekil 4.** Önerilen YSA modeli (Suggested ANN model)

Model oluşturulduktan sonra test verileri kullanılarak test edilmiştir. Test sonuçları modelin yüksek performansa sahip ve hata oranın oldukça düşük olduğunu ortaya koymuştur. Test edilen modelin deney sonuçları 4. bölümde verilmiştir.

## 4. BULGULAR VE TARTIŞMA (FINDINGS AND DISCUSSION)

Model belirlendikten ve ayrıntıları verildikten sonra performansı ölçülerek sonuçlar analiz edilmiştir.

**Deneysel Sonuçları** (Experimental Results)

Önerilen ağ yapılarının tahmin performanslarını ölçmek için literatürde yaygın olarak Ortalama Mutlak Yüzde Hata-MAPE (Eş.9), Ortalama Karesel Hata-MSE (Eş.10) ve Mutlak Değişim Yüzdesi-$R^2$ (Eş.11) formülleri kullanılmaktadır. Burada, Eşitlik 9-10'dan düşük; Eşitlik 11'den ise yüksek bir değer elde edilmesi tercih edilmektedir [33,35-37].

$$MAPE = \frac{1}{T} \sum \left| \frac{y_t - \hat{y}_t}{y_t} \right| \times 100 \quad (9) \qquad MSE = \frac{1}{T} \sum_{t=1}^{T} \left( y_t - \hat{y}_t \right)^2 \quad (10) \qquad R^2 = 1 - \left( \frac{\sum_t (y_t - \hat{y}_t)^2}{\sum_t (\hat{y}_t)^2} \right) \quad (11)$$

($y_t$ = Gerçek değerler, $\hat{y}_t$ = Tahmin değerleri, T = Tahmin sayısı)

Model (**8-10-10-1**), 2018 yılının Ocak ayına ait gerçek veriler kullanılarak test edilmiştir. Modelin tahmin performansı, Eşitlik 9-11 arasındaki formüller kullanılarak ölçülmüştür. Deney sonuçları Çizelge 5'te verilmiştir.

**Çizelge 5.** Deney sonuçları (The results of the experiment)

| No | Gerçek Değer | Tahmin Değeri | MAPE | MSE | $R^2$ |
|---|---|---|---|---|---|
| 1 | 0,81481 | 0,80399 | 0,59647 | 0,00012 | 99,40353 |
| 2 | 0,77778 | 0,78001 | 0,12556 | 0,00000 | 99,87444 |
| 3 | 0,73148 | 0,72856 | 0,16873 | 0,00001 | 99,83127 |
| 4 | 1,00000 | 1,00420 | 0,21000 | 0,00002 | 99,79000 |
| 5 | 0,64815 | 0,63916 | 0,54535 | 0,00008 | 99,45465 |
| 6 | 0,75000 | 0,75560 | 0,32000 | 0,00003 | 99,68000 |
| 7 | 0,86111 | 0,89687 | 1,92137 | 0,00128 | 98,07863 |
| 8 | 0,69444 | 0,71025 | 0,93279 | 0,00025 | 99,06721 |

| No | Gerçek Değer | Tahmin Değeri | MAPE | MSE | $R^2$ |
|---|---|---|---|---|---|
| 9 | 0,77778 | 0,77580 | 0,11125 | 0,00000 | 99,88875 |
| 10 | 0,88889 | 0,83520 | 2,84235 | 0,00288 | 97,15765 |
| 11 | 0,70370 | 0,74510 | 2,42978 | 0,00171 | 97,57022 |
| 12 | 0,99074 | 0,96580 | 1,25284 | 0,00062 | 98,74716 |
| 13 | 0,78704 | 0,78450 | 0,14197 | 0,00001 | 99,85803 |
| 14 | 0,67593 | 0,66890 | 0,41923 | 0,00005 | 99,58077 |
| 15 | 0,75000 | 0,74890 | 0,06286 | 0,00000 | 99,93714 |
| 16 | 0,85185 | 0,86540 | 0,73160 | 0,00018 | 99,26840 |
| 17 | 0,71296 | 0,70010 | 0,75092 | 0,00017 | 99,24908 |
| 18 | 0,75926 | 0,74590 | 0,75937 | 0,00018 | 99,24063 |
| 19 | 0,88889 | 0,88000 | 0,47059 | 0,00008 | 99,52941 |
| 20 | 0,70370 | 0,71200 | 0,48696 | 0,00007 | 99,51304 |
| 21 | 0,87037 | 0,87630 | 0,31703 | 0,00004 | 99,68297 |
| 22 | 0,78704 | 0,79540 | 0,46798 | 0,00007 | 99,53202 |
| 23 | 0,00000 | 0,00500 | 0,50000 | 0,00002 | 99,50000 |
| 24 | 0,96296 | 0,95986 | 0,15808 | 0,00001 | 99,84192 |
| 25 | 0,71296 | 0,68750 | 1,48649 | 0,00065 | 98,51351 |
| 26 | 0,76852 | 0,75841 | 0,57158 | 0,00010 | 99,42842 |
| 27 | 0,84259 | 0,83950 | 0,16784 | 0,00001 | 99,83216 |
| 28 | 0,78704 | 0,78100 | 0,33782 | 0,00004 | 99,66218 |
| 29 | 0,73148 | 0,73155 | 0,00396 | 0,00000 | 99,99604 |
| 30 | 0,78704 | 0,75200 | 1,96062 | 0,00123 | 98,03938 |
| 31 | 0,87037 | 0,87950 | 0,48812 | 0,00008 | 99,51188 |
| **Ortalama** | | | 0,701275 | 0,000322 | 99,29 |

Çizelge 5 incelendiğinde, çalışmanın ortalama 0,701275'lik bir oranla MAPE, 0,000322'lik bir oranla MSE ve 99,29'luk bir oranla $R^2$ değerleri ile yüksek performans oranları elde edilmiştir. Modelden elde edilen gerçek ve tahmin değerlerinin karşılaştırmalı gösterimi Şekil 5'te verilmiştir. Şekil 5 incelendiğinde, deney süreci içerisinde gerçek proje verileri kullanılarak model ile elde edilen tahmin sonuçlarının gerçek sonuçlarla büyük ölçüde örtüştüğü ve hata oranlarının da çok düşük (sıfıra yakın) olduğu görülmektedir.

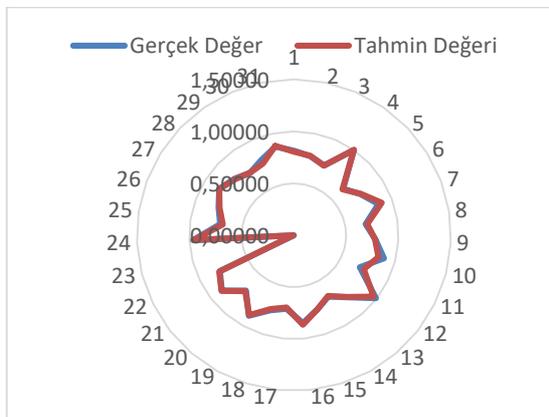

**Şekil 5.** Gerçek ve tahmin değerlerinin karşılaştırılması (Comparison of real and predicted values)

## 5. SONUÇ VE ÖNERİLER (CONCLUSION AND RECOMMENDATIONS)

Bu çalışmada, işletmelerde üretilen yemek miktarını veya personel talebini yapay sinir ağları kullanılarak tahmin eden bir model önerilmiştir. Önerilen model, gerçek verilerle test edilmiş ve sonuçları verilmiştir. Deney sonuçları, analizler ve karşılaştırmalar, önerilen YSA modelinin yüksek doğrulukta tahminler gerçekleştirebildiğini, dolayısıyla modelin performans bakımından oldukça iyi olduğunu açıkça ortaya koymuştur. Bu noktada, model belirlenirken gizli katmandaki nöron sayısının çok düşük veya çok yüksek belirlendiği durumlarda modelin performansının düştüğü, hata oranı da önemli düzeyde yükseldiği görülmüştür. Bu durum, modelin ezberlediğine veya öğrendiğini unuttuğuna işaret olarak gösterilebilir.

Bu bağlamda, YSA ile oluşturulan tahmin modellerinin etkili sonuçlar vermesine karşın, geleneksel metotlar ile bulunan deney sonuçları YSA'yı desteklemede yardımcı olarak kullanılabilir. Çalışmanın bir sonraki aşamasında, özellikle model oluşturulurken hibrid yaklaşımlar kullanılabilir. Çalışmanın, işletmeler veya kurumlar için gereğinden fazla harcanacak emeğin, paranın ve zamanın önüne geçmek veya en azından bu faktörleri minimize etmek için önemli bir uygulama örneği olduğu söylenebilir.